\documentclass[conference,10pt]{IEEEtran}
\usepackage{amsmath,amssymb,amsfonts}
\usepackage{graphicx}
\usepackage{textcomp}
\usepackage{amsmath}
\usepackage{booktabs} 
\usepackage{multirow}
\usepackage{hyperref}       % hyperlinks
\usepackage{url}            % simple URL typesetting
\usepackage{booktabs}       % professional-quality tables
\usepackage{amsfonts}       % blackboard math symbols
\usepackage{nicefrac}       % compact symbols for 1/2, etc.
\usepackage{microtype}      % microtypography
\usepackage{xcolor}         % colors
\usepackage{algorithm}
\usepackage{algpseudocode}
\usepackage{amsmath}
\usepackage{cite}
\usepackage{tikz}

\usepackage{orcidlink}
\usepackage{comment}
\usepackage{amssymb}
\usepackage{braket}
\usepackage{cite}
\usepackage{xcolor}
\def\BibTeX{{\rm B\kern-.05em{\sc i\kern-.025em b}\kern-.08em
    T\kern-.1667em\lower.7ex\hbox{E}\kern-.125emX}}
\begin{document}
\pagestyle{plain}

\title{QFAL: Quantum Federated Adversarial Learning}

%\author{\IEEEauthorblockN{Anonymous Authors}}
\author{\IEEEauthorblockN{Walid El Maouaki\textsuperscript{1}, Nouhaila Innan\textsuperscript{2,3}, Alberto Marchisio\textsuperscript{2,3}, Taoufik Said\textsuperscript{1}, \\ Mohamed Bennai\textsuperscript{1}, and Muhammad Shafique\textsuperscript{2,3}}
\IEEEauthorblockA{\textsuperscript{1}Quantum Physics and Spintronic Team, LPMC, Faculty of Sciences Ben M'sick,\\ Hassan II University of Casablanca,
Morocco\\
\textsuperscript{2}eBRAIN Lab, Division of Engineering, New York University Abu Dhabi (NYUAD), Abu Dhabi, UAE\\
\textsuperscript{3}Center for Quantum and Topological Systems (CQTS), NYUAD Research Institute, NYUAD, Abu Dhabi, UAE\\ walid.elmaouaki-etu@etu.univh2c.ma, nouhaila.innan@nyu.edu, alberto.marchisio@nyu.edu, taoufik.said@univh2c.ma, \\mohamed.bennai@univh2c.ma, muhammad.shafique@nyu.edu\\
}
%\vspace{-30pt}
}
\maketitle

\begin{abstract}

Quantum federated learning (QFL) merges the privacy advantages of federated systems with the computational potential of quantum neural networks (QNNs), yet its vulnerability to adversarial attacks remains poorly understood. This work pioneers the integration of adversarial training into QFL, proposing a robust framework, quantum federated adversarial learning (QFAL), where clients collaboratively defend against perturbations by combining local adversarial example generation with federated averaging (FedAvg). We systematically evaluate the interplay between three critical factors: client count (5, 10, 15), adversarial training coverage (0–100\%), and adversarial attack perturbation strength ($\epsilon=0.01-0.5$), using the MNIST dataset. 
Our experimental results show that while fewer clients often yield higher clean-data accuracy, larger federations can more effectively balance accuracy and robustness when partially adversarially trained.
Notably, even limited adversarial coverage (e.g., 20\%–50\%) can significantly improve resilience to moderate perturbations, though at the cost of reduced baseline performance. Conversely, full adversarial training (100\%) may regain high clean accuracy but is vulnerable under stronger attacks. These findings underscore an inherent trade-off between robust and standard objectives, which is further complicated by quantum-specific factors. We conclude that a carefully chosen combination of client count and adversarial coverage is critical for mitigating adversarial vulnerabilities in QFL. Moreover, we highlight opportunities for future research, including adaptive adversarial training schedules, more diverse quantum encoding schemes, and personalized defense strategies to further enhance the robustness–accuracy trade-off in real-world quantum federated environments.

\end{abstract}


\section{Introduction}

Federated Learning (FL) has emerged as an effective paradigm for collaborative training of Machine Learning (ML) models across distributed clients without requiring raw data to be centralized. This approach addresses critical concerns related to privacy, data sovereignty, and communication costs \cite{asad2023limitations}. Simultaneously, quantum computing has demonstrated the potential to accelerate certain computational tasks and enable new forms of ML architectures through quantum neural networks (QNNs) \cite{beer2020training,abbas2021power,innan2024lep,innan2025qnn,innan2024variational}. By combining FL's collaborative training mechanism with quantum computing's ability to encode and process information in qubits, quantum federated learning (QFL) has the potential to harness benefits from both domains, pointing to a promising avenue for next-generation distributed intelligence \cite{innan2024fedqnn,chehimi2022quantum}.
However, ML models, both classical and quantum, are susceptible to adversarial attacks~\cite{finlayson2019adversarial, marchisio2023adversarial, chattopadhyay2024defending, guesmi2024dap, el2024advqunn, el2024robqunns, lu2020quantum}. Minor, carefully crafted perturbations to input data can significantly degrade a model's performance at inference time. In the context of QFL, the distributed nature of training and the unique structure of QNNs create new challenges in both detecting and countering adversarial attacks. Existing adversarial defense strategies in FL often focus on classical deep networks, leaving open questions about how best to incorporate adversarial robustness techniques in a QFL setting~\cite{zhang2023delving}. At the same time, partial adoption of adversarial training by only a subset of federated clients introduces unexplored dimensions for mitigating attack vectors, especially when dealing with the limited qubit resources and data encoding restrictions inherent to many quantum devices.

In this work, we propose a novel QFL framework that integrates local adversarial training within each client's QNN. By systematically varying the number of clients, the fraction of clients that perform adversarial training, and the perturbation strengths of the adversarial examples, we aim to evaluate the framework's robustness and scalability. Our results illustrate the trade-offs between clean-data accuracy and defense against adversarial examples in quantum federated scenarios, providing guidance for practitioners and researchers interested in designing secure, privacy-conscious QML solutions. 

Our work offers several key contributions to quantum QFL and adversarial robustness:

\begin{itemize}
    \item We propose QFAL, a novel quantum federated adversarial learning framework that seamlessly integrates adversarial training into quantum federated learning, specifically designed to counteract the unique vulnerabilities inherent to quantum federated learning. 

    \item We develop a tailored local adversarial training protocol that leverages PGD-based adversarial example generation within each quantum client, enabling a dynamic balance between training on clean and perturbed data. This method enhances resilience against malicious inputs during inference, and introduces client-level heterogeneity in defense strategies. 

    \item We conduct a comprehensive empirical evaluation of QFAL by systematically varying client counts, adversarial training coverage, and perturbation strengths. Our experiments reveal that even partial adversarial training (e.g., 20\% coverage in a 5-client system) can boost robustness by over 12\% compared to baseline models, while larger federations (10–15 clients) yield a more favorable trade-off between clean-data accuracy and adversarial resilience.
\end{itemize}

\textbf{Paper Organization:} Section~\ref{sec:background} discusses the background and related work in QML, QFL, and security. Section~\ref{sec:methodology} presents the details of our proposed QFAL framework. Section~\ref{sec:results} reports the experimental results. Section~\ref{sec:discussion} summarizes the key findings of this study. Section~\ref{sec:conclusion} concludes the paper and discusses future directions.

\section{Background and Related Work}
\label{sec:background}
\subsection{Quantum Machine Learning}

QML is an interdisciplinary field combining quantum computing with classical ML paradigms to enhance computational efficiency and tackle complex learning tasks across various domains \cite{biamonte2017quantum,schuld2015introduction}. At the core of QML are QNNs, which leverage parameterized quantum circuits (PQCs) as trainable models for various learning problems.

QNNs extend classical neural networks by utilizing quantum circuits as function approximators. A QNN consists of multiple quantum layers, each composed of quantum gates parameterized by trainable variables $\theta$. The evolution of the quantum state follows:
\begin{equation}
    |\psi(\theta)\rangle = U_L(\theta_L) \dots U_2(\theta_2) U_1(\theta_1) |0\rangle^{\otimes n},
\end{equation}
where each unitary transformation $U_i(\theta_i)$ corresponds to a quantum layer applied sequentially to an initial quantum state. After processing the input state, the final quantum state $|\psi(\theta)\rangle$ is measured to extract predictions.

Encoding classical data into quantum states is a fundamental step in QNNs. Given a classical input vector $x \in \mathbb{R}^d$, the goal is to map it to a quantum state $|\psi(x)\rangle$ in Hilbert space. Several encoding techniques exist, including basis encoding, angle encoding, and amplitude encoding.

QNNs are trained by minimizing a loss function, typically in a supervised learning setting. The loss function is computed after measurements are performed on the quantum state. A commonly used loss function is the Mean Squared Error (MSE):
\begin{equation}
    L(\theta) = \frac{1}{N} \sum_{i=1}^{N} (y_i - f(\theta; x_i))^2,
\end{equation}
where $y_i$ represents the true label, and $f(\theta; x_i)$ is the predicted output obtained from quantum measurements.

Since quantum measurements collapse the wavefunction, multiple circuit evaluations (shots) are required to estimate the expectation value accurately. The function \( f(\theta; x) \) is defined as the expectation value of an observable $\hat{O}$, often chosen as a Pauli operator (e.g., $\hat{Z}$):
\begin{equation}
    f(\theta; x) = \langle \psi(\theta) | \hat{O} | \psi(\theta) \rangle.
\end{equation}
However, due to shot noise, this estimation introduces statistical variance that can impact training stability, particularly in gradient-based optimization.

To optimize the parameters $\theta$, classical gradient-based techniques are employed. The parameter-shift rule, widely used in quantum circuits, efficiently extracts gradients for certain parameterized gates:
\begin{equation}
    \frac{\partial L(\theta)}{\partial \theta_j} = \frac{L(\theta + \frac{\pi}{2} e_j) - L(\theta - \frac{\pi}{2} e_j)}{2},
\end{equation}
where $e_j$ is a unit vector in the direction of the $j$-th parameter. This technique is valid when the parameterized quantum gates are of the form $e^{-i\theta G}$, where $G$ is a Hermitian generator (such as Pauli operators). Once the gradients are extracted using the parameter-shift rule, optimization algorithms such as Adam adaptively update the parameters to minimize the loss function. 
%More details about Adam, but we can keep them as a comment for now until we finish the paper, and based on the space, we can add

\begin{comment}
The Adam optimizer, which combines momentum and adaptive learning rates, updates parameters using the following equations:
\begin{equation}
    m_t = \beta_1 m_{t-1} + (1 - \beta_1) \nabla_{\theta} L(\theta), \quad v_t = \beta_2 v_{t-1} + (1 - \beta_2) \nabla_{\theta} L(\theta)^2,
\end{equation}
\begin{equation}
    \hat{m}_t = \frac{m_t}{1 - \beta_1^t}, \quad \hat{v}_t = \frac{v_t}{1 - \beta_2^t},
\end{equation}
\begin{equation}
    \theta_{t+1} = \theta_t - \frac{\eta}{\sqrt{\hat{v}_t} + \epsilon} \hat{m}_t,
\end{equation}
where $m_t$ and $v_t$ represent the first and second moment estimates (mean and variance of gradients), $\beta_1$ and $\beta_2$ are decay rates for momentum and variance, $\eta$ is the learning rate, and $\epsilon$ is a small constant to prevent division by zero.
\end{comment}
\subsection{QFL}

QFL extends classical FL to the quantum domain, enabling multiple quantum clients to collaboratively train a global quantum model while preserving data privacy and adhering to quantum constraints such as the no-cloning theorem. In a standard QFL setup, \(K\) quantum clients, each holding a local quantum dataset, train individual QML models. Instead of sharing raw quantum data, the global model is updated through an aggregation process that integrates updates from all clients. Each training round consists of the following steps (see Algorithm \ref{alg:qfl}), each client trains its local QML model using its available quantum data, computes classical representations of quantum gradients or parameter updates (e.g., via the parameter-shift rule), and transmits them to a central server. The central server aggregates the updates using a weighted averaging mechanism:
\begin{equation}
    \theta^{(t+1)} = \sum_{k=1}^{K} w_k \theta_k^{(t)},
    \label{updates}
\end{equation}
where \( \theta_k^{(t)} \) denotes the parameters of the \( k \)-th QNNs at iteration \( t \), \( w_k = \frac{|D_k|}{\sum_{i=1}^{K} |D_i|} \) is the weight assigned to each client  \( k \), proportional to its dataset size, typically proportional to dataset size or computational resources, and \( \theta^{(t+1)} \) is the updated global model. The weights satisfy the normalization condition:
\begin{equation}
    \sum_{k=1}^{K} w_k = 1.
\end{equation}
The updated global model is redistributed to all clients, and this process repeats until convergence based on a predefined stopping criterion. This ensures privacy preservation while facilitating collaborative quantum model training across decentralized clients.

\begin{algorithm}
\caption{QFL}
\label{alg:qfl}
\begin{algorithmic}[1]
\Require Number of clients \( K \), number of rounds \( T \), local epochs \( E \), learning rate \( \eta \)
\State Initialize global model \( \theta^{(0)} \)
\For{each round \( t = 0, 1, \dots, T-1 \)}
    \For{each client \( k \in \{1, \dots, K\} \) \textbf{in parallel}}
        \State Download global model \( \theta^{(t)} \)
        \For{each local epoch \( e = 1, \dots, E \)}
            \State Compute quantum gradients using the parameter-shift rule
            \State Update local model: \( \theta_k^{(t)} \gets \theta_k^{(t)} - \eta \nabla_{\theta} \mathcal{L}_k \)
        \EndFor
        \State Send local update \( \theta_k^{(t)} \) to the server
    \EndFor
    \State Server aggregates updates: \( \theta^{(t+1)} = \sum_{k=1}^{K} w_k \theta_k^{(t)} \)
\EndFor
\State \textbf{Return} final global model \( \theta^{(T)} \)
\end{algorithmic}
\end{algorithm}

\subsection{Adversarial Attacks and Defenses in Classical ML}

%(Alberto)

\begin{figure}[ht]
  \centering
  \includegraphics[width=\linewidth]{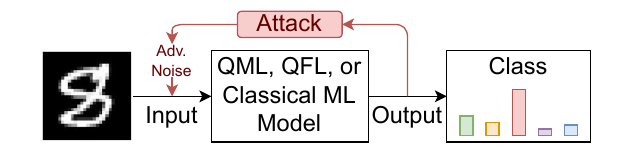}
  \caption{A simplified representation of an adversarial attack scenario, where the perturbations added to the image alter the classification process. A similar approach for the attack can be applied to various models, including classical ML, QML, and QFL.}
  \label{fig:attacks_QFAL}
\end{figure}

ML models, particularly deep neural networks, are highly vulnerable to adversarial examples—small, imperceptible perturbations to input data that lead to incorrect model predictions \cite{szegedy2013intriguing}. As simple representation of an adversarial attack scenario is shown in Fig.~\ref{fig:attacks_QFAL}. This phenomenon raises significant concerns in security-sensitive applications such as finance, healthcare, and autonomous systems. Addressing these vulnerabilities requires a thorough understanding of adversarial attack strategies and corresponding defense mechanisms.

\subsubsection{Projected Gradient Descent (PGD) Attack}
One of the most effective and widely used adversarial attack methods is the Projected Gradient Descent (PGD) attack \cite{madry2017towards}. PGD iteratively crafts adversarial perturbations by maximizing the classification loss of a given model under a bounded constraint. 

Given a classifier parameterized by $\theta$, the PGD attack iteratively perturbs an input sample to maximize the classification loss while ensuring the perturbation remains within a predefined bound. This process involves adjusting the input in the direction of the loss gradient, guided by a predefined step size. The adversarial modifications are constrained within a bounded region around the original input to ensure they remain within an acceptable perturbation range. The iterative nature of PGD makes it one of the strongest first-order attacks, as it effectively refines the perturbation to maximize its impact on the model's prediction.

Formally, given a classifier $f_\theta$ parameterized by $\theta$, an input sample $x$, and its corresponding label $y$, PGD generates an adversarial example $x'$ using:

\subsubsection{Adversarial Training}
Adversarial training is a widely used defense mechanism that enhances model robustness by training on adversarially perturbed samples \cite{madry2017towards}. The core idea is to modify the training process such that the model learns to correctly classify adversarial examples.
By incorporating adversarial examples into the training process, adversarial training increases model resilience against such attacks.

However, adversarial training introduces computational overhead due to the iterative nature of attack generation during training and may lead to slight performance degradation on clean samples \cite{tsipras2018robustness}. Despite these trade-offs, adversarial training remains a cornerstone defense method for securing ML models against adversarial threats.

In the context of QFL, integrating adversarial training with quantum models presents additional challenges due to limited quantum resources and noise-induced perturbations. Our work builds upon these principles to develop robust federated quantum models resistant to adversarial manipulations.

\subsection{Related work}

Early QFL research demonstrated the feasibility of federated training on hybrid quantum-classical models, achieving comparable accuracy with faster training \cite{chen2021federated}. QFL has since been extended to variational quantum algorithms, optimizing variational quantum circuits while preserving data privacy \cite{huang2022quantum}. The federated quantum neural network framework further advanced QFL by integrating quantum machine learning with classical federated learning, showing high accuracy across genomics and healthcare datasets \cite{innanqfl}. Additionally, QFL has been explored in spiking neural networks, where the FL-QDSNNs framework leverages a dynamic threshold mechanism to enhance learning efficiency while ensuring scalability and privacy in distributed quantum environments \cite{innanqsnn}.

Privacy-preserving QFL protocols leveraging blind quantum computation \cite{li2021quantum} have enhanced security in distributed quantum systems. Other developments include quantum fuzzy federated learning for improved robustness in non-IID settings \cite{qu2024quantum} and federated quantum natural gradient descent for reducing communication overhead and accelerating convergence \cite{qi2024federated}. Quantum differential privacy has been explored to mitigate model inversion attacks and data leakage \cite{rofougaran2024federated}, while QFL integration with fully homomorphic encryption has improved security and robustness in multimodal federated environments \cite{dutta2024mqfl,dutta2024federated}.

Despite advancements in QFL, adversarial robustness remains underexplored, with limited research addressing vulnerabilities to malicious interventions. In security-sensitive applications such as autonomous vehicles and cybersecurity, FL models are susceptible to adversarial attacks, including data poisoning and Byzantine attacks. For instance, quantum-behaved particle swarm optimization has been employed to enhance model resilience against poisoning attacks in federated autonomous vehicle networks \cite{qi2023optimizing}, while quantum-inspired federated averaging techniques have been introduced for cyber-attack detection using spatio-temporal attention networks \cite{subramanian2024hybrid}.

In the QFL context, Byzantine resilience has been studied, demonstrating that classical Byzantine-tolerant algorithms can be adapted to quantum federated settings \cite{xia2021defending}. However, the impact of adversarial evasion attacks—exploiting quantum-specific properties such as superposition, entanglement, and measurement uncertainty—remains largely unexplored. While classical adversarial methods like FGSM \cite{goodfellow2014explaining} and PGD \cite{madry2017towards} are well studied, their applicability to quantum models within a federated framework is unclear.
Motivated by this critical shortfall, our work aims to rigorously analyze the impact of evasion attacks in a QFL context and propose novel defense mechanisms that enhance the adversarial robustness of QFL models.

\section{QFAL Methodology}
\label{sec:methodology}

\begin{figure*}[htb]
    \centering
    \includegraphics[width=1\linewidth]{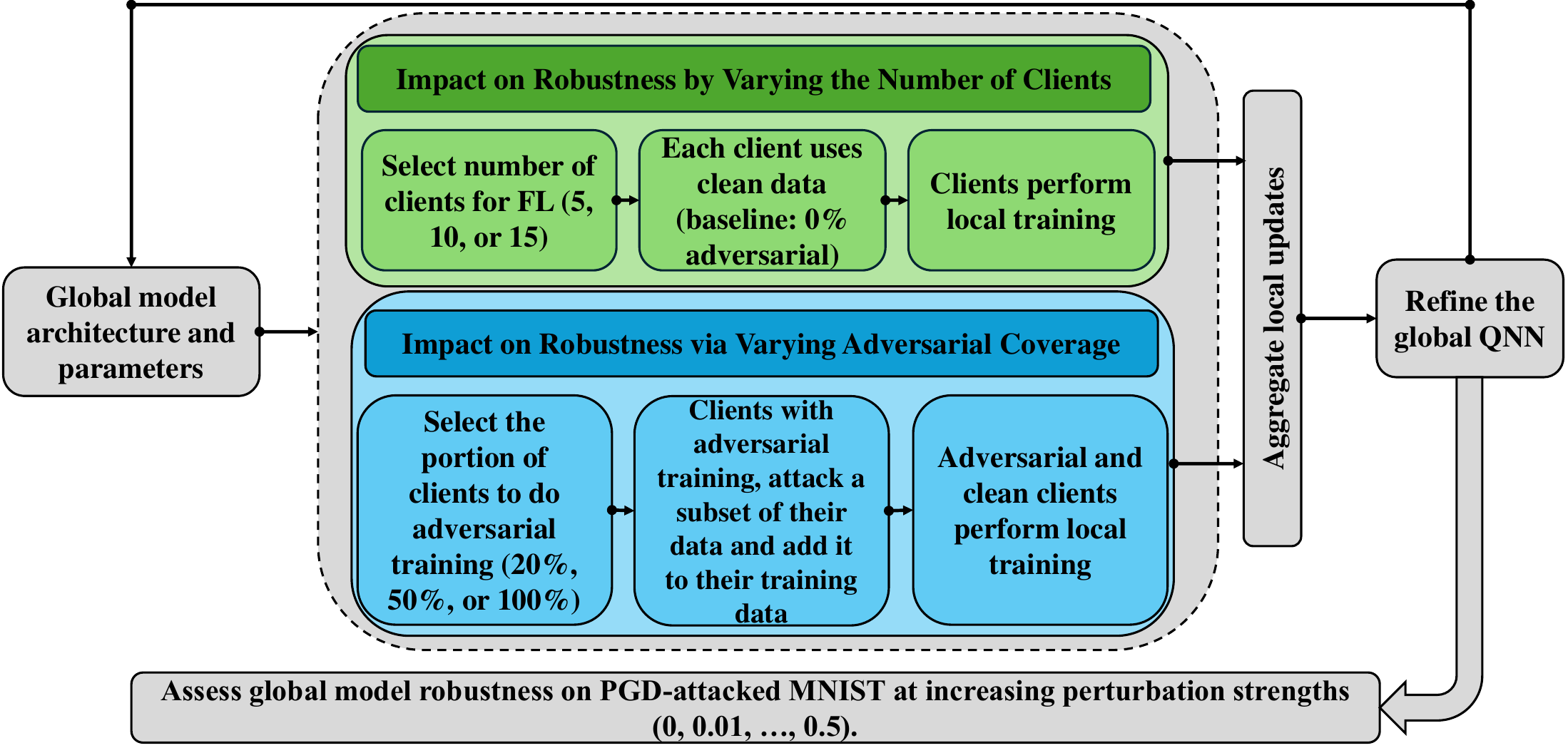}
    \caption{Overview of our proposed QFL framework for adversarially robust quantum training. Each client maintains a local QNN, trained on its unique subset of MNIST data and (optionally) augmented with adversarial examples generated via PGD. After a set number of local epochs, clients upload their updated parameters to a central server, where they are aggregated using FedAvg.}
    \label{methodology_fig}
\end{figure*}

Our QFAL framework integrates a QNN within an FL setup, employing adversarial training with a quantum-enhanced defense algorithm to improve robustness. The evaluation assesses accuracy, resilience, and privacy. A schematic is shown in Fig.~\ref{methodology_fig}, and the algorithm is detailed in Algorithm~\ref{alg:qfal}.

\subsection{QNN Architecture}

Each participating client maintains a local QNN trained on an independently drawn, identically distributed (IID) subset of the MNIST dataset (restricted to digits 0, 1, and 2, resized to \(8\times8\) pixels for quantum state preparation). The quantum model is based on PQCs, which apply trainable quantum gates to process classical inputs.
The QNN employs amplitude encoding, where classical input features are embedded into the amplitudes of a quantum state, allowing for an efficient representation of high-dimensional data. Given a normalized input vector \( x = (x_1, x_2, \dots, x_d) \), amplitude encoding maps it into a quantum state as:
\begin{equation}
    |\psi(x)\rangle = \sum_{i=1}^{d} x_i |i\rangle,
\end{equation}
where each coefficient \( x_i \) is directly encoded in the amplitude of the computational basis state \( |i\rangle \). This approach enables an exponential compression of classical data into quantum states but requires quantum operations that maintain normalization constraints.
The trainable quantum layers perform feature transformations through entangling gates and rotation operators, with final measurements in the Pauli-Z basis extracting class predictions.

\begin{figure}[htpb]
    \centering
    \vspace{-0.2cm}
    \includegraphics[width=0.7\linewidth]{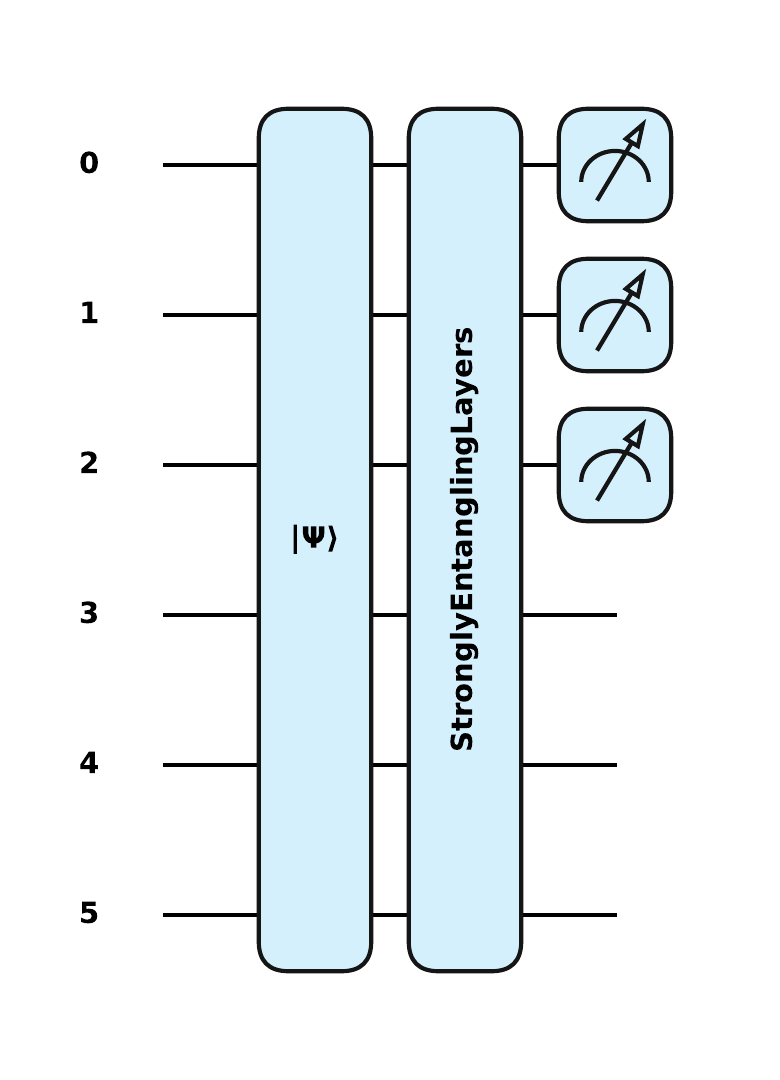}
    \vspace{-1cm}
    \caption{QNN circuit utilizing amplitude embedding for state preparation, where classical input features are encoded into a normalized quantum state, followed by strongly entangling layers that apply trainable parametrized rotations and controlled entanglement to enhance feature representation. Measurement in the Pauli-Z basis on selected qubits provides expectation values corresponding to class predictions, enabling quantum-assisted classification.}
    \label{QNN}
\end{figure}

\subsection{QFL Training Protocol}

Federated training in QFAL follows a standard FL procedure using the FedAvg aggregation scheme. At each communication round \( t \), a subset of \( K \) participating clients receives the current global model \( \theta^{(t)} \) and trains their local QNNs using their private datasets. Each client \( k \) updates its model parameters \( \theta_k^{(t)} \) by minimizing a local loss function \( \mathcal{L}_k \), typically computed over its dataset \( D_k \):
\begin{equation}
    \theta_k^{(t+1)} = \theta_k^{(t)} - \eta \nabla_{\theta} \mathcal{L}_k(\theta_k^{(t)}),
\end{equation}
where \( \eta \) represents the learning rate. After local training, each client transmits its updated model parameters to the central server. The server performs aggregation using the FedAvg approach, computing the new global model as a weighted sum of the client updates as per equation \ref{updates}.
% \begin{equation}
%     \theta^{(t+1)} = \sum_{k=1}^{K} w_k \theta_k^{(t+1)},
% \end{equation}
The updated global parameters \( \theta^{(t+1)} \) are then distributed back to all participating clients, and this process iterates for multiple rounds until convergence. The iterative optimization aims to minimize a global objective function defined as:
\begin{equation}
    \min_{\theta} \sum_{k=1}^{K} \frac{|D_k|}{|D|} \mathcal{L}_k(\theta),
\end{equation}
where \( |D| = \sum_{k=1}^{K} |D_k| \) is the total dataset size across all clients. This decentralized approach enables collaborative training of the global QNN while preserving client data privacy. The process continues until a predefined stopping criterion is met, such as a convergence threshold on the loss function or a fixed number of communication rounds.

\subsection{Adversarial Training for Robust QFL}
A core innovation in QFAL is the incorporation of local adversarial training, which significantly improves robustness against adversarial perturbations.
During local training, selected clients generate adversarial examples using the PGD attack, formulated as:
\begin{equation}
    x_{t+1} = \Pi_{\mathcal{B}_\epsilon(x)} \left( x_t + \alpha \cdot \text{sign} \left( \nabla_x \mathcal{L}(f_\theta(x_t), y) \right) \right),
\end{equation}
where \( \mathcal{L} \) is the model loss function, \( \alpha \) is the step size, \( \mathcal{B}_\epsilon(x) \) is the adversarial perturbation constraint, and \( \Pi \) ensures the perturbation remains within the permissible range.
To enhance robustness, participating clients mix clean samples and adversarially perturbed samples within each mini-batch during training. This adversarial training follows a min-max optimization paradigm:
\begin{equation}
    \min_\theta \mathbb{E}_{(x,y) \sim \mathcal{D}} \max_{\| \delta \| \leq \epsilon} \mathcal{L}(f_\theta(x + \delta), y),
\end{equation}
where \( \delta \) represents adversarial perturbations, \( \epsilon\) controls perturbation strength. Clients performing adversarial training update their QNN parameters accordingly before sending them to the server.

\subsection{Testing and Evaluation}

To systematically assess the effectiveness and scalability of QFAL, we evaluate model performance across three key factors: the number of participating clients, the extent of adversarial coverage, and the perturbation strength of adversarial attacks. First, we analyze the impact of varying the number of clients by considering scenarios with 5, 10, and 15 clients, allowing us to measure how increasing client participation affects model accuracy and robustness. Second, we assess the effect of adversarial training coverage by varying the fraction of clients performing adversarial training, considering cases where 0\%, 20\%, 50\%, and 100\% of clients incorporate adversarial examples into their local training. This helps determine the trade-off between adversarial robustness and federated training stability. Finally, we evaluate the model's resilience to adversarial perturbations by progressively increasing the PGD attack strength parameter \( \epsilon \), providing insight into how the global model performs under different levels of adversarial pressure.

For optimization, we employ the Adam optimizer, which combines momentum-based updates with adaptive learning rates to improve convergence stability. The first and second moment estimates of the gradients are updated as:

\begin{equation}
    m_t = \beta_1 m_{t-1} + (1 - \beta_1) \nabla_{\theta} L(\theta), \quad v_t = \beta_2 v_{t-1} + (1 - \beta_2) \nabla_{\theta} L(\theta)^2,
\end{equation}

where \( m_t \) and \( v_t \) represent the first and second moment estimates of the gradients, respectively, and \( \beta_1 \) and \( \beta_2 \) are decay rates that control momentum and variance adaptation. To correct for bias in these estimates, they are normalized as:

\begin{equation}
    \hat{m}_t = \frac{m_t}{1 - \beta_1^t}, \quad \hat{v}_t = \frac{v_t}{1 - \beta_2^t}.
\end{equation}

The model parameters are then updated using:

\begin{equation}
    \theta_{t+1} = \theta_t - \frac{\eta}{\sqrt{\hat{v}_t} + \epsilon} \hat{m}_t,
\end{equation}
where \( \eta \) is the learning rate, and \( \epsilon \) is a small constant to prevent numerical instability. 

%======================Pseudo code===============================
\begin{algorithm}[ht]
\caption{QFAL}
\label{alg:qfal}
\begin{algorithmic}[1]
\Require 
  $D$: MNIST (classes 0,1,2) resized to $8\times8$,
  $nC \in \{5,10,15\}$,
  Coverage $\in \{0\%,\,20\%,\,50\%,\,100\%\}$,
  \textsc{FedAvg}($\cdot$)

\State Split $D$ IID into $\{D_1,\dots,D_{nC}\}$
\For{\textbf{each} cov $\in \{0\%,20\%,50\%,100\%\}$}
    \If{\text{cov} = 0\%}
        \State $\text{Global}_0 \gets \text{initModel}(); \text{Rounds}=50$ \Comment{Initialize a fresh global model}
    \Else
        \State $\text{Global}_0 \gets \text{loadBaseline}(); \text{Rounds}=20$ \Comment{Use the model from previous coverage = 0\% as a starting point}
    \EndIf
    
    \For{$r = 1$ \textbf{to} Rounds}
        \For{$i = 1$ \textbf{to} $nC$}
            \State $\text{Local}_i \gets \text{Global}_{r-1}$
            \If{$i$ in coverage portion}
                \State $\text{W}_i \gets \text{localTrainAdv}(\text{Local}_i,D_i)$
            \Else
                \State $\text{W}_i \gets \text{localTrainClean}(\text{Local}_i,D_i)$
            \EndIf
        \EndFor
        \State $\text{Global}_r \gets \textsc{FedAvg}(\{\text{W}_i\}_{i=1}^{nC})$
    \EndFor
    
    \State \textsc{Evaluate}($\text{Global}_r$, \text{CleanTest}, \text{AdvTest}) \Comment{Test on clean and adversarial data at various \(\epsilon\)}
    \State \textsc{SaveModel}($\text{Global}_r,\text{cov}$)
\EndFor
\end{algorithmic}
\end{algorithm}

\section{Results}
\label{sec:results}
\subsection{Experimental Setup}
In this study, we focus on a subset of the MNIST dataset containing only the digits 0, 1, and 2. To reduce the input dimension for our QNN, each $28\times28$ image is downsampled to an $8\times8$ resolution. This enables amplitude encoding into a 6-qubit circuit for the QNN. Once processed, the dataset is split into multiple parts and distributed among clients in an IID manner, ensuring each client receives a balanced portion of the data.
Each client independently maintains and updates a QNN composed of multiple layers of parameterized quantum gates, mixed with entanglement operations. In our approach, we employ two layers of strongly entangled PQCs following the data encoding step. The final layer, known as the measurement layer, produces class probabilities, which are used to compute a cross-entropy loss during training. We use the Adam optimizer with a learning rate of 0.01, and each client runs local updates for four epochs per federated training round on mini-batches of size 64.
To coordinate learning across clients, we employ a standard federated averaging approach. After completing their local training, clients send their updated parameters to a central server. The server aggregates these parameters by averaging them to form a global model, which is then redistributed to all clients.

In terms of robustness, clients designated for adversarial training generate adversarial examples using a PGD-based method, parameterized by $\epsilon=0.1$, ten iterations, and a step size of $\alpha=0.01$. During each local update, the mini-batch is split in half: one portion is trained on clean images, while the other portion is trained on adversarially perturbed images. This adversarial training framework is then scaled across different proportions of clients ($0 \%, 20 \%, 50 \%, 100 \%$), reflecting the fraction of the population actively mitigating attacks.
We investigate the influence of adversarial training under varying coverage (0\%, 20\%, 50\%, and 100\% of clients) and different client counts (5, 10, and 15), to determine how the federated system scales. In the baseline case (0\% adversarial coverage), all clients train exclusively on clean data for 50 rounds. For the subsequent scenarios (20\%, 50\%, and 100\% coverage), we initialize local models from the global model obtained in the baseline and continue training for an additional 20 rounds, allowing us to quantify the trade-off between clean data accuracy and adversarial robustness, while also revealing the effect of partial versus complete adversarial training coverage in a QFL environment.
%allowing us to isolate the effect of incremental adversarial coverage.

We further examine how the system performs against varying adversarial strengths at inference. Specifically, we evaluate the final global model by generating test-time adversarial examples with $\epsilon$ values ranging from 0.01 to 0.5. In each case, ten iterations of PGD are used, and the step size is set to $\alpha=\epsilon / 10$. Through these configurations, we obtain a comprehensive view of the model's resilience as the threat level escalates. The results are reported in Tables \ref{tab:5__10_15clients}, \ref{tab:5_clients}, \ref{tab:10_clients}, and \ref{tab:15_clients}, where each cell shows the accuracy (\%) of its respective experiment.

\subsection{Convergence and Performance Under Adversarial Coverage}

\begin{figure}
    \centering
    \includegraphics[width=\linewidth]{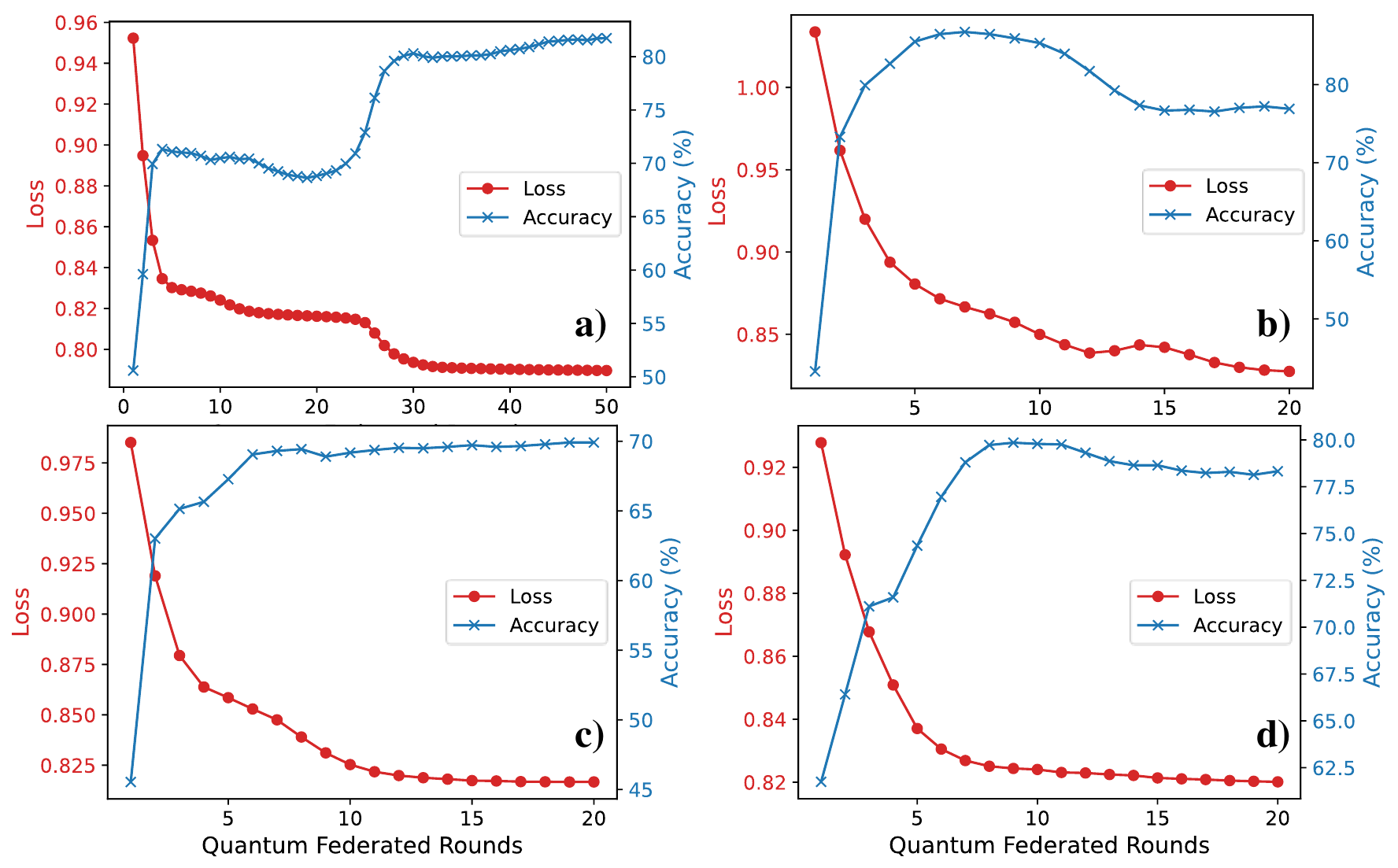}
    \caption{Convergence of global test loss and progression of test accuracy for 5 clients. (a) Convergence on clean data across 50 rounds, (b) Convergence with 20\% adversarial-data coverage across 20 rounds, (c) Convergence with 50\% adversarial-data coverage across 20 rounds, and (d) Convergence with 100\% adversarial-data coverage across 20 rounds.}
    \label{fig:5clients}
\end{figure}

\begin{figure}
    \centering
    \includegraphics[width=\linewidth]{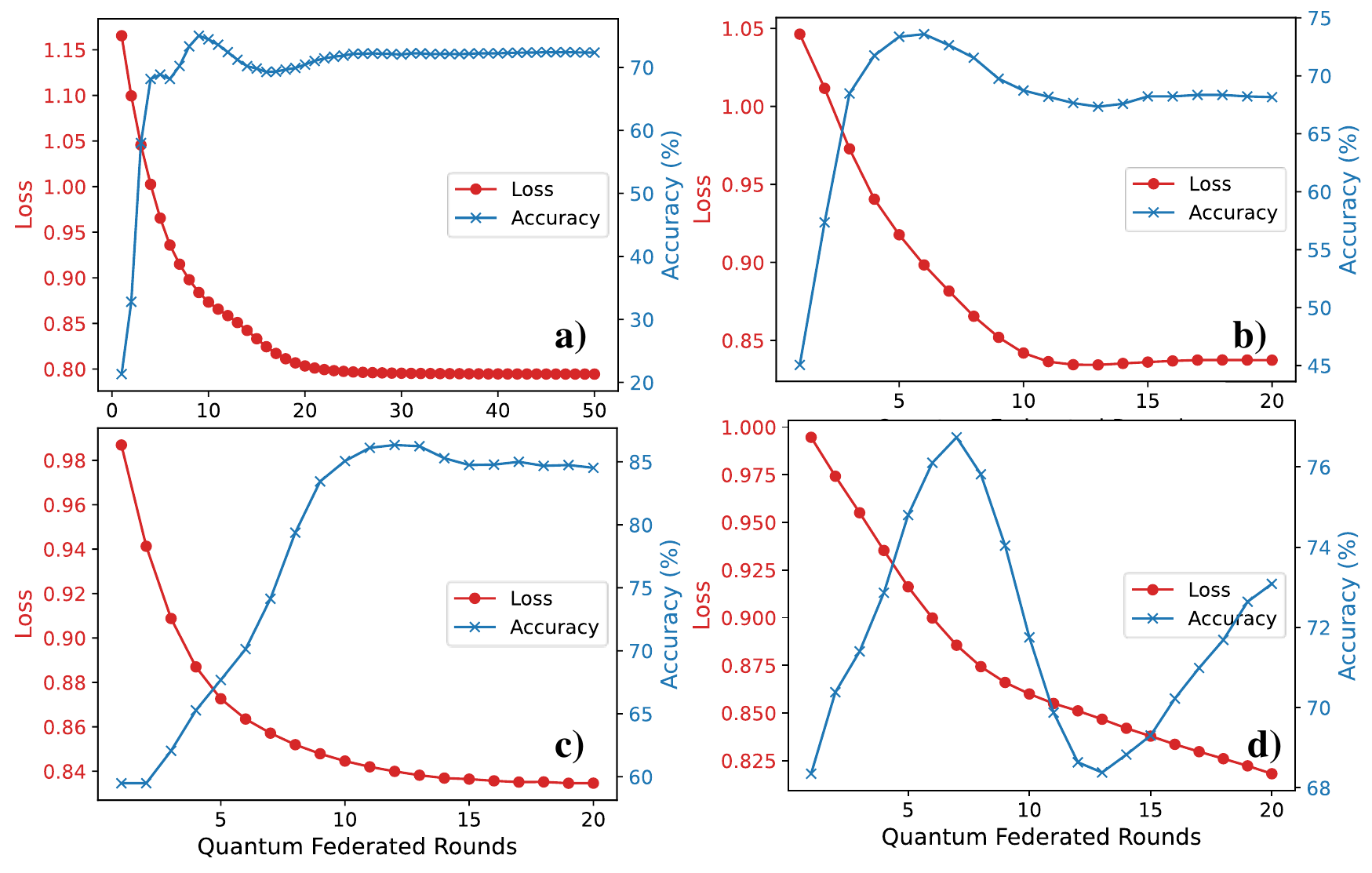}
    \caption{Convergence of global test loss and progression of test accuracy for 10 clients. (a) Convergence on clean data across 50 rounds, (b) Convergence with 20\% adversarial-data coverage across 20 rounds, (c) Convergence with 50\% adversarial-data coverage across 20 rounds, and (d) Convergence with 100\% adversarial-data coverage across 20 rounds.}
    \label{fig:10clients}
\end{figure}

\begin{figure}
    \centering
    \includegraphics[width=\linewidth]{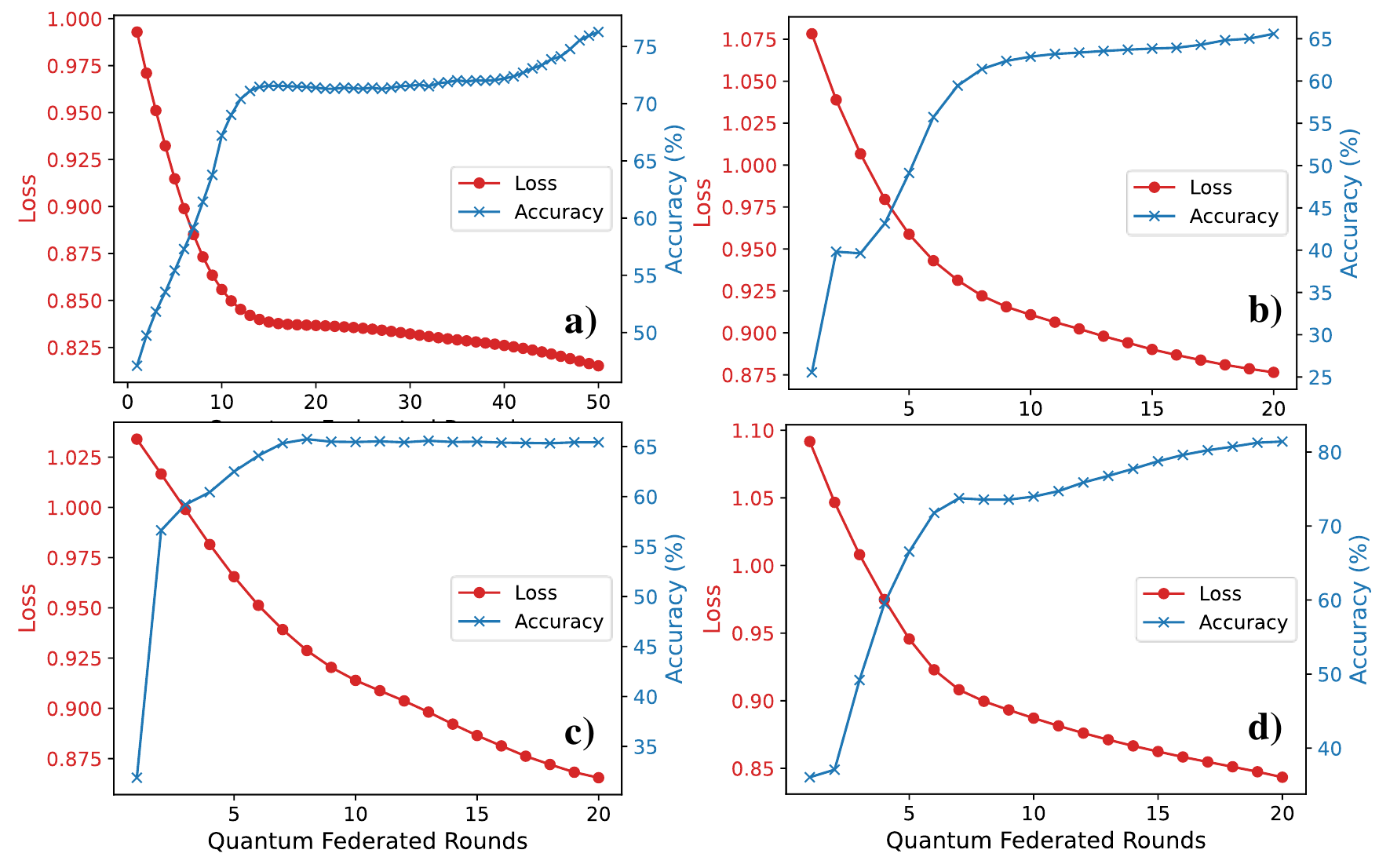}
    \caption{Convergence of global test loss and progression of test accuracy for 15 clients. (a) Convergence on clean data across 50 rounds, (b) Convergence with 20\% adversarial-data coverage across 20 rounds, (c) Convergence with 50\% adversarial-data coverage across 20 rounds, and (d) Convergence with 100\% adversarial-data coverage across 20 rounds.}
    \label{fig:15clients}
\end{figure}

Figures \ref{fig:5clients}, \ref{fig:10clients} and \ref{fig:15clients} illustrate the convergence of global test loss and test accuracy over 50 QFL rounds for 5, 10, and 15 clients in clean training, and over 20 rounds for adversarial training. Compare performance under both clean conditions (0\% adversarial data) and varying levels of adversarial data coverage (20\%, 50\%, and 100\%). In the clean case, test loss decreases (e.g., from $\approx 0.96$ to $\approx 0.80$ for 5 clients) and accuracy rises steadily (plateauing near $81\%$ for 5 clients), confirming stable convergence in non-adversarial conditions. As adversarial data coverage increases, final accuracy drops and test loss remains higher, but the system continues to converge across all client configurations, demonstrating the underlying robustness of QFL to adversarial presence, albeit with diminished ultimate performance relative to the purely clean baseline.

\subsection{Impact of Client Count on Robustness}
Table \ref{tab:5__10_15clients} presents a comparison of global model accuracy under adversarial attacks with varying perturbation strengths ($\epsilon$) for systems comprising 5,10 , and 15 clients, all trained without adversarial defense ($0 \%$ coverage). The results indicate that clean accuracy ($\epsilon=0$) is highest for the system with 5 clients (81.73\%), followed by 15 clients (76.26\%) and 10 clients (72.35\%). As the perturbation strength increases, all configurations experience a sharp decline in accuracy. Notably, at $\epsilon=0.1$, the system with 10 clients demonstrates marginally higher robustness ($37.88 \%$) compared to those with 5 and 15 clients, which achieve $32.13 \%$ and $32.19 \%$, respectively. However, for $\epsilon \geq 0.2$, accuracy collapses to near-zero across all configurations, indicating a high vulnerability to strong adversarial perturbations.

\subsection{Effect of Adversarial Training Coverage}
Table \ref{tab:5_clients} presents the performance of a system with 5 clients under different levels of adversarial defense coverage. The baseline configuration (0\% coverage) achieves the highest clean accuracy (81.73\%) but exhibits poor robustness, with accuracy dropping to 32.13\% at $\epsilon = 0.1$
%, and collapsing entirely for stronger adversarial perturbations (ϵ ≥ 0.3).
. Introducing 20\% coverage enhances robustness at $\epsilon = 0.1$ to 44.61\%, representing a 12.48\% improvement over the baseline, while maintaining moderate clean accuracy (76.90\%)
%albeit at a cost to clean accuracy (76.90%)
. Increasing coverage to 50\% further strengthens robustness at $\epsilon = 0.2$ (9.53\%); however, this comes at the cost of a reduction in clean accuracy (69.91\%). A fully adversarially trained model (100\% coverage) strikes a balance, achieving a clean accuracy of 78.33\% while maintaining moderate robustness, with accuracy reaching 39.80\% at $\epsilon = 0.1$.

Table \ref{tab:10_clients} presents the performance of a system with 10 clients under different levels of adversarial defense coverage. The configuration with 50\% coverage achieves the highest clean accuracy (84.52\%) and exhibits the best robustness at $\epsilon = 0.05$, reaching an accuracy of 64.57\%. Increasing the coverage to 100\% results in the highest robustness at $\epsilon = 0.1$, achieving an accuracy of 42.39\%, which represents an improvement of $4.51\%$ over the baseline. Meanwhile, the 20\% coverage setting demonstrates the highest resilience at $\epsilon = 0.2$, with an accuracy of 9.37\%, but at the cost of reduced clean accuracy (68.16\%).

Table \ref{tab:15_clients} presents the performance of a system with 15 clients under varying levels of adversarial defense coverage. The configuration with 20\% coverage provides optimal robustness at $\epsilon = 0.1$ and $\epsilon = 0.2$, achieving accuracies of 40.90\% and 8.29\%, respectively. In contrast, the 100\% coverage setting maximizes clean accuracy (81.41\%) and exhibits the highest robustness at $\epsilon = 0.05$ (61.84\%), but its performance deteriorates at higher perturbation strengths. The baseline configuration, lacking adversarial defense, experiences the steepest accuracy decline, dropping to 32.19\% at $\epsilon = 0.1$.

Our analysis reveals inherent trade-offs between adversarial robustness and clean accuracy in QFL systems. Adversarial training enhances resilience to attacks (e.g., 50\% coverage in 10 clients improves $\epsilon = 0.05$ accuracy by $+7.4\%$) but often reduces clean accuracy ($-8.17\%$ versus baseline). Higher adversarial coverage ($50-100\%$) amplifies robustness to moderate perturbations ($\epsilon \leq 0.1$) at the expense of clean performance, while systems with $10-15$ clients strike the most effective balance between these metrics. Notably, all configurations collapse catastrophically ($\leq 0.5\%$ accuracy) for $\epsilon \geq 0.3$, highlighting the diminishing returns of adversarial training against stronger attacks. Accuracy declines nonlinearly with increasing $\epsilon$: for instance, in a 5-client system with 20\% coverage, accuracy significantly drops from 74.10\% ($\epsilon = 0.01$) to 3.46\% ($\epsilon = 0.2$). Partial adversarial coverage (20-50\%) proves optimal for practical deployments, particularly in intermediate-sized federations (10 clients), where robustness gains outweigh clean accuracy losses. These findings underscore the need to tailor adversarial training coverage to both client count and anticipated threat severity.

%========================impact of varying the number of clients on robustness =====================================

\begin{table}[!ht]
\centering
\caption{Global model accuracy (\%) when varying the number of QFL clients (5, 10, 15).}
\label{tab:5__10_15clients}
\begin{tabular}{l|ccccccc}
\toprule
\multirow{2}{*}{\textbf{Num of Clients}} & \multicolumn{7}{c}{\textbf{Test Accuracy (\%)}} \\ 
\cline{2-8}
& \(\epsilon = 0\) & $0.01$ & $0.05$ & $0.1$ & $0.2$ & $0.3$ & $0.5$ \\
\midrule
5 clients  &  \textbf{81.73} & \textbf{77.09} & \textbf{58.63} & 32.13 & 0.13 & 0 & 0 \\
10 clients  &  72.35 & 68.86 & 57.17 & \textbf{37.88} & \textbf{2.93} & 0 & 0 \\
15 clients  &  76.26 & 71.40 & 54.85 & 32.19 & 0.35 & 0 & 0 \\
\bottomrule
\end{tabular}
\end{table}

%================= TABLE 1: 5 Clients =================
\begin{table}[!ht]
\centering
\caption{Global model accuracy (\%) for \textbf{5 clients}. We evaluate on the clean test set (\(\epsilon=0\)) and on adversarially perturbed test sets with different \(\epsilon\).}
\label{tab:5_clients}
\begin{tabular}{l|ccccccc}
\toprule
\multirow{2}{*}{\textbf{Adv Training}} & \multicolumn{7}{c}{\textbf{Test Accuracy (\%)}} \\ 
\cline{2-8}
& \(\epsilon = 0\) & $0.01$ & $0.05$ & $0.1$ & $0.2$ & $0.3$ & $0.5$ \\
\midrule
0\% (Baseline)  &  \textbf{81.73} & \textbf{77.09} & 58.63 & 32.13 & 0.13 & 0 & 0 \\
20\%            &  76.90 & 74.10 & \textbf{62.54} & \textbf{44.61} & 3.46 & 0 & 0 \\
50\%            &  69.91 & 67.75 & 58.47 & 41.40 & \textbf{9.53} & \textbf{0.13} & 0 \\
100\%           &  78.33 & 75.47 & 59.93 & 39.08 & 4.42 & 0 & 0 \\
\bottomrule
\end{tabular}
\end{table}

%================= TABLE 2: 10 Clients  =================
\begin{table}[!ht]
\centering
\caption{Global model accuracy (\%) for \textbf{10 clients}. We evaluate on the clean test set (\(\epsilon=0\)) and on adversarially perturbed test sets with different \(\epsilon\).}
\label{tab:10_clients}
\begin{tabular}{l|ccccccc}
\toprule
\multirow{2}{*}{\textbf{Adv Training}} & \multicolumn{7}{c}{\textbf{Test Accuracy (\%)}} \\ 
\cline{2-8}
& \(\epsilon = 0\) & $0.01$ & $0.05$ & $0.1$ & $0.2$ & $0.3$ & $0.5$ \\
\midrule
0\% (Baseline)  &  72.35 & 68.86 & 57.17 & 37.88 & 2.93 & 0 & 0 \\
20\%            &  68.16 & 65.40 & 54.94 & 37.40 & \textbf{9.37} & \textbf{0.06} & 0 \\
50\%            &  \textbf{84.52} & \textbf{81.34} & \textbf{64.57} & 38.99 & 2.38 & 0 & 0 \\
100\%           &  73.09 & 69.65 & 58.28 & \textbf{42.39} & 6.55 & 0 & 0 \\
\bottomrule
\end{tabular}
\end{table}

%================= TABLE 3: 15 Clients=================
\begin{table}[!ht]
\centering
\caption{Global model accuracy (\%) for \textbf{15 clients}. We evaluate on the clean test set (\(\epsilon=0\)) and on adversarially perturbed test sets with different \(\epsilon\).}
\label{tab:15_clients}
\begin{tabular}{l|ccccccc}
\toprule
\multirow{2}{*}{\textbf{Adv Training}} & \multicolumn{7}{c}{\textbf{Test Accuracy (\%)}} \\ 
\cline{2-8}
& \(\epsilon = 0\) & $0.01$ & $0.05$ & $0.1$ & $0.2$ & $0.3$ & $0.5$ \\
\midrule
0\% (Baseline)  &  76.26 & 71.40 & 54.85 & 32.19 & 0.35 & 0 & 0 \\
20\%            &  65.59 & 63.11 & 55.20 & \textbf{40.90} & \textbf{8.29} & 0 & 0 \\
50\%            &  65.43 & 64.16 & 57.07 & 39.18 & 4.93 & 0 & 0 \\
100\%           &  \textbf{81.41} & \textbf{77.57} & \textbf{61.84} & 35.84 & 0.44 & 0 & 0 \\
\bottomrule
\end{tabular}
\end{table}

%%%%%%%%%%%%%%%%%%%%%%%%%%%%%%%%%%%%%%%%%%%%%%%
\begin{figure}
    \centering
    \includegraphics[width=\linewidth]{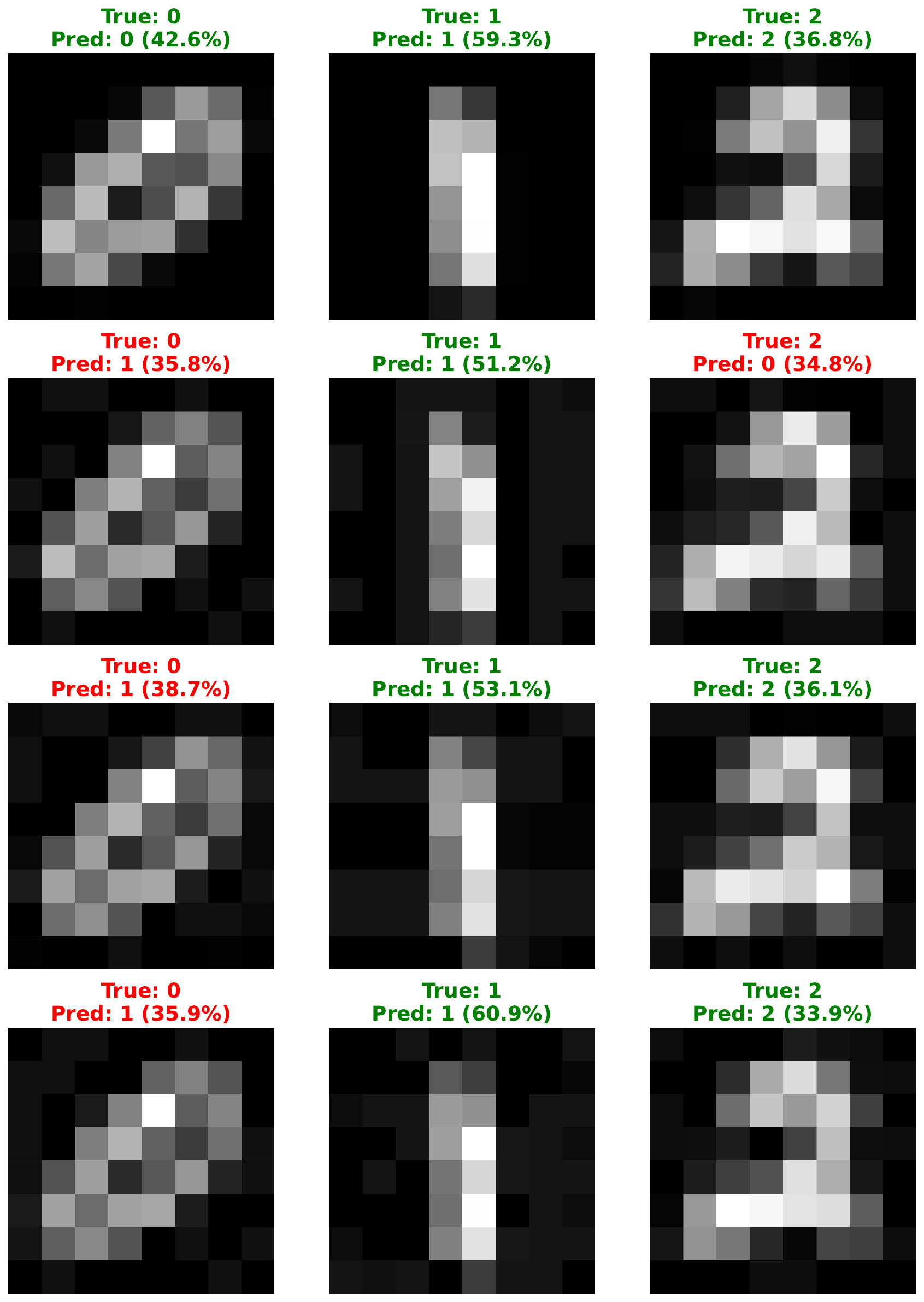}
    \caption{MNIST digit predictions for digits 0, 1, and 2 under varying levels of adversarial robustness. The first row displays clean images with no adversarial training (0\%). The second row presents images attacked using PGD ($\epsilon=0.1$) from the 0\% adversarial training model. The third row shows PGD-attacked images from the 20\% adversarial training model, while the fourth row contains PGD-attacked images from the 50\% adversarial training model. Green borders indicate correct predictions, whereas red borders highlight misclassifications, with confidence scores provided for each prediction.}
    \label{attack_imgs}
\end{figure}
%%%%%%%%%%%%%%%%%%%%%%%%%%%%%%%%%%%%%%%%%%%%%%%

Fig. \ref{attack_imgs} presents the predictions for MNIST digits 0, 1, and 2 in a federated learning setup with 5 clients, comparing clean images to those perturbed by a PGD attack ($\epsilon=0.1$) across models trained with $0 \%, 20 \%$, and 50\% adversarial training coverage. In row 1, the baseline ($0 \%$ coverage) model correctly classifies all digits with moderate confidence (e.g., $42.6 \%$ for digit $0,59.3 \%$ for digit 1 , and $36.8 \%$ for digit 2). In row 2, under attack, the $0 \%$ model misclassifies digit 0 as 1 ($35.8 \%$ confidence) and digit 2 as 0 (34.8\%), though digit 1 remains correct (51.2\%). With 20\% adversarial training (row 3), digit 0 is still misclassified (38.7\%), but digits 1 and 2 are predicted correctly ($53.1 \%$ and $36.1 \%$ respectively). Finally, row 4 shows that the $50 \%$ coverage model, while still misclassifying digit 0 ($35.9 \%$), correctly identifies digits 1 and 2 with improved confidence ($60.9 \%$ and $33.9 \%$). Green borders denote correct predictions, while red borders indicate misclassifications, highlighting that increased adversarial training coverage enhances robustness against PGD attacks in this federated framework.

\section{Discussion}
\label{sec:discussion}

The integration of adversarial training into QFL introduces nuanced trade-offs between model accuracy, robustness, and scalability. Our experiments reveal several insights into the dynamics of QFL under adversarial conditions.
A central tension emerges between maintaining high clean accuracy and achieving adversarial robustness. While adversarial training enhances resilience to perturbations (e.g., 20\% coverage in 5-client systems improves accuracy at $\epsilon=0.1$ by 12.48\%), it often degrades performance on clean data (Table 1). This aligns with classical FL observations, where robustness improvements typically come at the cost of generalization \cite{goodfellow2014explaining}. 

The number of clients significantly influences system performance. Smaller federations (5 clients) achieve superior baseline accuracy (81.73\%, Table 1), likely due to reduced parameter divergence during FedAvg aggregation. However, larger federations (10–15 clients) demonstrate greater potential for balancing accuracy and robustness through adversarial training. For example, 10-client systems with 50\% coverage retain moderate robustness ($\epsilon=0.1$: 38.99\%) without sacrificing clean accuracy (84.52\%), while 15-client systems with 20\% coverage achieve the highest robustness at $\epsilon=0.2$ (8.29\%, Table 3). This implies that distributed adversarial training across more clients—even at lower coverage—can enhance generalization by diversifying adversarial examples during aggregation.

Despite its benefits, adversarial training in QFL struggles against high-intensity attacks ($\epsilon \geq 0.3$), where accuracy collapses to near-zero levels across all configurations. This contrasts with classical FL, where adversarial defenses often maintain non-trivial robustness even under extreme perturbations \cite{madry2017towards}. We hypothesize that amplitude encoding—while efficient for quantum state preparation—may amplify vulnerability to input perturbations, as small adversarial changes in pixel values propagate nonlinearly through the quantum circuit. Future work could explore alternative encoding schemes (e.g., feature map embeddings) or employ complex quantum circuits designs to counteract adversarial perturbations \cite{maouaki2024designing}.

%hybrid quantum-classical architectures to mitigate this issue.
Interestingly, partial adversarial coverage ($20–50\%$) frequently outperforms full coverage ($100\%$) in balancing accuracy and robustness in certain $\epsilon$ ranges. For example, $20\%$ coverage in $5$-client systems achieves $44.61\%$ accuracy at $\epsilon=0.1$ (Table 1), whereas $100\%$ coverage yields only $39.08\%$. This suggests that heterogeneous training—where some clients focus on clean data and others on adversarial examples—preserves diverse feature representations, preventing overfitting to adversarial patterns. This finding echoes recent classical FL studies advocating for task-specialized clients \cite{yu2020heterogeneous}, but extends the concept to quantum domains.
% this indicate that a mix of standard and robust objectives may help the model generalize more flexibly across both clean and perturbed inputs. 

These findings suggest that practitioners may tune both the number of clients and the fraction of adversarially trained clients to balance clean accuracy with robustness, depending on the anticipated threat model. Smaller FL networks can excel in benign conditions but may require a higher fraction of adversarial training when moderate-to-high perturbations are likely. Conversely, larger networks might benefit from carefully chosen subsets of adversarially trained clients, which can improve robustness without sacrificing too much baseline performance.
In general, the observed dynamics indicate that there is a necessity for more refined strategies that tailor adversarial training levels to the distribution of data across clients. Future work could explore adaptive adversarial training schedules (e.g., high-risk clients train with stronger perturbations), hybrid defensive techniques (e.g quantum noise injection \cite{du2021quantum} or dynamic adversarial client selection strategies) to reduce the impact of balancing robustness and accuracy. Additionally, expanding the scope to stronger or more diverse attacks could reveal new insights into quantum federated defenses for real-world scenarios.

\section{Conclusion}
\label{sec:conclusion}

In this work, we introduced QFAL, a framework that integrates adversarial training into quantum federated learning to address the unique vulnerabilities of quantum neural networks in distributed environments. Our investigation, conducted on a tailored MNIST subset, demonstrated that even limited adversarial participation among clients can markedly enhance resilience against moderate attacks, while also uncovering a nontrivial trade-off between clean-data accuracy and robustness. Notably, our experiments indicate that the optimal balance depends on both the size of the client federation and the fraction of clients engaging in adversarial training.

Beyond validating the feasibility of adversarial defenses in QFL, our results highlight key challenges such as the collapse of accuracy under strong perturbations and the sensitivity introduced by amplitude encoding. These insights suggest that further refinement of quantum data encoding methods, as well as adaptive and heterogeneous training strategies, may be necessary to fully harness the potential of quantum federated systems in adversarial settings.

Looking forward, expanding our framework to incorporate dynamic adversarial scheduling, exploring alternative quantum architectures, and testing under a broader array of attack models will be crucial for advancing secure distributed quantum machine learning. We anticipate that the QFAL framework and our experimental results will inspire further exploration into resilient quantum federated systems, paving the way for the creation of dependable quantum-enhanced solutions in practical applications.

%\section*{Acknowledgment}

\section*{Acknowledgment}
This work was supported in parts by the NYUAD Center for Quantum and Topological Systems (CQTS), funded by Tamkeen under the NYUAD Research Institute grant CG008, and the NYUAD Center for Cyber Security (CCS), funded by Tamkeen under the NYUAD Research Institute Award G1104.
\bibliographystyle{IEEEtran}

\bibliography{refs}

\end{document}